\documentclass{article}

\usepackage[dvipsnames,svgnames]{xcolor}

\PassOptionsToPackage{sort&compress}{natbib}

\usepackage[final]{corl_2026} % Uncomment for the camera-ready ``final'' version.
\makeatletter
\renewcommand{\@noticestring}{10th Conference on Robot Learning (CoRL 2026), Austin, Texas, USA.}
\makeatother

\usepackage{graphicx}
\usepackage{amsmath}
\usepackage{amssymb}
\usepackage{booktabs}
\usepackage{bbm}
\usepackage{multirow}
\usepackage{siunitx}
\usepackage{url}
\usepackage{caption}
\usepackage{adjustbox}
\usepackage[normalem]{ulem}
\usepackage{bm}
\usepackage{microtype}
\usepackage{wrapfig}
\usepackage{enumitem}
\usepackage{placeins}
\setlist[itemize]{topsep=1pt,itemsep=0pt,parsep=0pt,partopsep=0pt,leftmargin=1.4em}
\graphicspath{{../rss_submission/}}

\definecolor{myred}{RGB}{203,47,57}
\definecolor{myblue}{RGB}{21,102,151}
\definecolor{myorange}{RGB}{249,149,51}
\definecolor{pmcolor}{RGB}{70,70,70}
\definecolor{mygreen}{RGB}{0,205,0}
\definecolor{mybrown}{RGB}{139,69,19}
\definecolor{FR}{HTML}{EDE9F5}
\definecolor{FE}{HTML}{844380}
\definecolor{S}{HTML}{B13D2D}
\definecolor{H}{HTML}{BADCFF}
\definecolor{FL}{HTML}{E48B50}

\newcommand{\pmcolor}[1]{\textcolor{pmcolor}{#1}}

\newcommand{\ci}[2]{\,{\tiny{\pmcolor{[#1,#2]}}}}
\newcommand{\paragraphc}[1]{\vspace{0.13em}\noindent\textbf{#1.}}

\title{How to Peel with a Knife: Aligning Fine-Grained Manipulation with Human Preference}

\author{
  Toru Lin\thanks{Equal contribution.} \quad Shuying Deng\rlap{$^{*}$} \quad Zhao-Heng Yin \quad Pieter Abbeel \quad Jitendra Malik \\
  \\
  University of California, Berkeley \\[0.6em]
  \href{https://toruowo.github.io/peel/}{\texttt{https://toruowo.github.io/peel}}
}

\hypersetup{
  pdftitle={How to Peel with a Knife: Aligning Fine-Grained Manipulation with Human Preference},
  pdfauthor={Toru Lin, Shuying Deng, Zhao-Heng Yin, Pieter Abbeel, Jitendra Malik}
}

\begin{document}
\maketitle

\vspace{-15pt}
\begingroup
  \centering
  \captionsetup{hypcap=false}
  \includegraphics[width=0.96\textwidth]{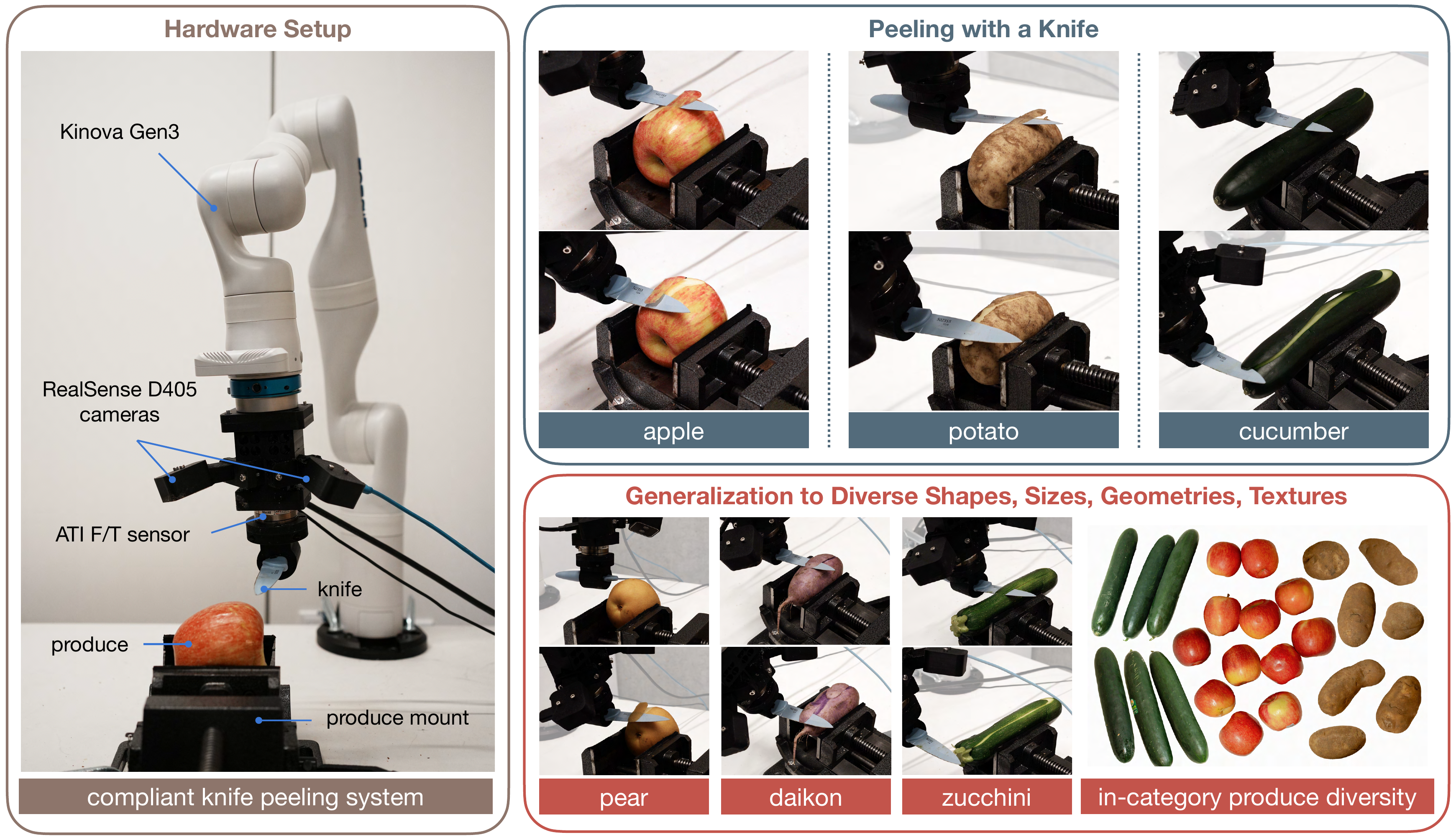}
  \captionof{figure}{\small{\textbf{An overview of our system setup and learned peeling policies.}
    We use a 7-DoF Kinova Gen3 arm with impedance control. A custom designed mount holding a knife is attached to the tool end. Two wrist cameras are attached to the tool end and pointing towards the knife and produce. We collect data on three types of produce, train peeling policies that zero-shot transfer to three selected held-out produce types, and finetune the policies to improve peel quality as evaluated by a human annotator.}}
  \label{fig:teaser}
  \par
\endgroup

\begin{abstract}
Many essential manipulation tasks -- such as food preparation, surgery, and craftsmanship -- remain difficult for autonomous robots. These tasks are characterized not only by contact-rich, force-sensitive dynamics, but also by their ``implicit'' success criteria: unlike pick-and-place, task quality in these domains is continuous and subjective (e.g. how well a potato is peeled), making quantitative evaluation and reward design difficult. We present a learning framework for such tasks, using peeling with a knife as a representative example. Our approach follows a two-stage pipeline: first, we learn a robust initial policy via force-aware data collection and imitation learning, enabling generalization across object variations; second, we refine the policy through preference-based finetuning using a learned reward model that combines quantitative task metrics with qualitative human feedback, aligning policy behavior with human-evaluated task quality. Using only 50--200 peeling trajectories, our system achieves 100\% same-produce success on cucumbers, apples, and potatoes, with performance improving by up to 40\% through preference-based finetuning. Remarkably, policies trained on one produce category also transfer zero-shot to selected held-out produce categories, achieving 50\% success on zucchini, 90\% on pear, and 80\% on daikon radish.
\end{abstract}


\vspace{-5pt}
\section{Introduction}
\vspace{-5pt}

Many essential manipulation tasks -- such as food preparation, surgery, and craftsmanship -- remain challenging for autonomous robots despite recent progress in learning-based robotic manipulation~\cite{fang2023anygrasp,lin2025sim,wang2022dexgraspnet,lum2024dextrah,intelligence2025pi}. The fundamental bottlenecks lie in two coupled aspects: (1) \emph{quantity} -- collecting high-quality demonstrations at scale is difficult when contact is delicate and failures can damage the object; (2) \emph{quality} -- task success is often continuous, subjective, and difficult to specify mathematically, making it hard to evaluate learning outcomes and optimize for objectives that match human judgment.

In this work, we study how to address these bottlenecks through the lens of \textit{peeling with a knife}, a representative task in this problem class. Unlike using a peeler -- where the blade depth and contact angle are mechanically regulated -- peeling with a knife requires precise force control under unstable blade-surface contact, accurate real-time tracking of complex geometries, and generalization across variances of natural produce. Removing this constraint makes the task harder and more general; a knife is also preferred where a peeler is imprecise, e.g. for small patches of skin or inset blemishes. Furthermore, success is inherently difficult to quantify: beyond removing the skin, performance is also judged by the cleanliness, evenness, and efficiency of the cut.
Prior works address these challenges only partially: model-based controllers are brittle to modeling errors, calibration drift, and object variation; learning-based methods tend to require large-scale data collection that is expensive or infeasible; and evaluation is often reduced to fixed quantitative metrics that poorly align with human judgments of quality, limiting real-world applicability.

We present a learning framework that addresses both the \emph{quantity} and \emph{quality} bottlenecks by separating skill acquisition from refinement. The key insight is to first learn a force-aware policy that stays within desirable contact patterns from demonstrations, then refine the policy using preference-based rewards derived from quantitative and qualitative human evaluation.
Our approach follows a two-stage pipeline. First, we initialize a peeling policy using force-aware imitation learning, providing a strong baseline that generalizes across object variations. Second, we learn a reward model from human feedback and use it to finetune the policy, aligning its behavior with human-evaluated peel quality.
We evaluate our method on real-world knife-based peeling across multiple produce categories, including cucumbers, apples, and potatoes. Using only 50--200 trajectories, our approach achieves 100\% success on same-produce evaluations, with performance improving by up to 40\% after preference-based finetuning. Remarkably, policies trained on one produce category also generalize zero-shot to selected held-out produce categories with different geometry, compliance, and texture.

Our primary contributions are as follows. \textbf{First, a two-stage learning framework for aligning fine-grained manipulation with human-evaluated task quality.} We separate learning the basic motion and contact patterns from refining human-evaluated task quality -- through combining compliant data collection, force-aware imitation learning, and preference-based finetuning. \textbf{Second, hybrid preference rewards for peel quality.} We combine local quantitative labels for peel thickness with trajectory-level qualitative labels for overall peel quality, learn a reward model from these labels, and use the learned model to finetune policies. \textbf{Lastly, real-world evaluation of data-efficient transfer.} We show that visual, proprioceptive, and force sensing enable the learning of knife-peeling policies from 50--200 trajectories, including same-produce evaluations and zero-shot transfer to selected held-out produce categories.

Our results show that, for knife-based peeling, a policy initialized from demonstrations that already capture desirable motion and contact patterns can be further improved by preference-guided refinement -- enabling precise and adaptive contact-rich manipulation from limited real-world experience.
While our experiments are limited to knife-based peeling, we hypothesize that the same pipeline may extend to other manipulation tasks where human-evaluated quality is important.

\vspace{-5pt}
\section{Related Work}\vspace{-5pt}

\paragraphc{Learning manipulation from human preference} Modeling and learning from human preference is a long-standing problem in machine learning, with existing work spanning reinforcement learning~\cite{christiano2017deep,wirth2017survey} and supervised learning~\cite{furnkranz2003pairwise}. Recently, this problem has gained renewed attention through applications to large language models~\cite{chang2024survey}, and a growing body of work in robot learning~\cite{ibarz2018reward,hejna2023contrastive,hejna2023few,lee2021pebble,chen2025fdpp}. However, existing robotics work largely focuses on simple tasks in simulation or highly constrained real-world settings -- e.g. low-dimensional control, short-horizon manipulation, and binary success criteria -- where preference modeling and reward learning are relatively straightforward~\cite{christiano2017deep,ibarz2018reward,lee2021pebble,hejna2023few}.
Our goal is narrower than general preference learning: rather than modeling diverse preferences across users, we align a real-robot policy with a fixed, human-defined notion of task quality, specified by an expert annotator through a rubric that combines a quantitative thickness metric with a qualitative holistic judgment. We conduct this study on knife-based peeling, where quality is continuous and tightly coupled to subtle force-motion interactions.

\begin{figure}[t]
    \centering
    \includegraphics[width=\linewidth]{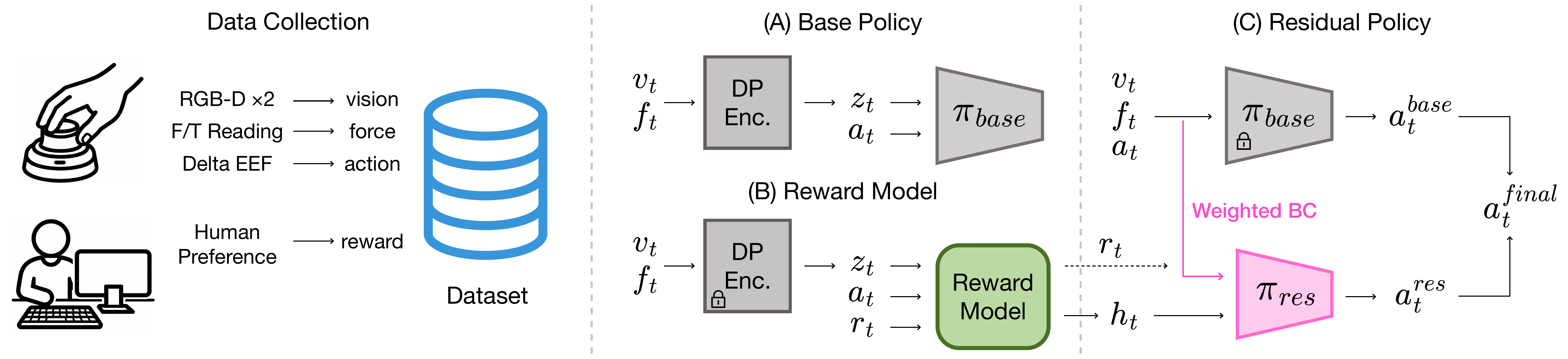}
    \caption{\textbf{An overview of our two-stage learning framework.} \emph{Data collection:} we teleoperate the robot and record two wrist RGB-D images, force-torque (F/T) readings, and delta end-effector (EEF) actions, which are annotated with human preference labels. \emph{(A) Base policy:} a diffusion policy (DP) encoder maps vision $v_t$ and force $f_t$ to a latent feature $z_t$, from which $\pi_{\mathrm{base}}$ predicts the action $a_t$. \emph{(B) Reward model:} with the DP encoder frozen (lock icon), the reward model is trained to regress the human-derived reward $r_t$ from $(z_t, a_t)$ and exposes a hidden representation $h_t$. \emph{(C) Residual policy:} with $\pi_{\mathrm{base}}$ frozen, $\pi_{\mathrm{res}}$ predicts a correction $a_t^{\mathrm{res}}$ from $z_t$, the base action $a_t^{\mathrm{base}}$, and $h_t$, and is trained with reward-weighted behavioral cloning (Weighted BC); the executed action is $a_t^{\mathrm{final}} = a_t^{\mathrm{base}} + a_t^{\mathrm{res}}$.}
    \label{fig:overview}
\end{figure}
\paragraphc{Peeling with a knife} We review adjacent tasks, including knife cutting and peeler-based peeling. Prior peeler-based systems have demonstrated meaningful peeling capability using model-based planning~\cite{watanabe2013cooking,dong2021food}, teleoperation data~\cite{xue2025reactive}, or scripted policies~\cite{chen2024vegetable,ye2024morpheus}. Our focus is complementary: knife-based peeling removes the mechanical constraint provided by a peeler and therefore requires precise regulation of force, blade angle, and depth along a continuously evolving contact surface to avoid slippage or breakage. Existing knife-cutting works~\cite{long2014force,mu2023dexterous,mu2019robotic,jamdagni2024robotic,yang2016vision,han2020vision,straivzys2020surfing} rely primarily on \uline{classical model-based approaches}, differing mainly in heuristics and analytical models of force dynamics, perception, and knife motion. While these methods demonstrate feasibility, they exhibit limited generalization due to sensitivity to modeling errors and perception noise. \uline{Learning-based approaches} remain scarce: RoboNinja \cite{xu2023roboninja} combines model-based control with a specialized differentiable cutting simulator. These limitations highlight the difficulty of collecting high-quality data and learning compliant policies for realistic knife-based peeling.

\paragraphc{Force-based manipulation} Knife-based peeling is inherently force-sensitive, motivating learning-based approaches for generalization. Prior work incorporates force in the observation space, the action space, or both~\cite{he2025foar,liu2025forcemimic,xue2025reactive,chen2025dexforce,hou2025adaptive}. To use force in observation space, existing works propose architectures to process signals from tactile or force-torque sensors, and add the encoded feature into observation vector. To use force in action space, a force-based controller such as impedance controller or admittance controller is often used to achieve compliant control. To tackle the challenge of data collection and learning compliant control, ACP~\cite{hou2025adaptive} and DexForce~\cite{chen2025dexforce} propose to collect data with kinesthetic teaching and recover the compliance parameters by estimating effective mass and inertia; but these estimations are largely hand-tuned and rule-based, limiting applicability to hard-to-model tasks like peeling. Other works collect data using specialized force-aware teleoperation systems~\cite{he2025foar,liu2025forcemimic}, which restrict accessibility. Simulation-based approaches for cutting~\cite{heiden2021disect,xu2023roboninja} and sim-to-real transfer~\cite{zhang2023efficient,zhang2024bridging} have shown promise, but remain difficult to scale beyond simple insertion or cutting tasks due to the complexity of modeling cutting dynamics in deformable, heterogeneous produce~\cite{yin2021modeling,sochacki2024towards}.

\vspace{-5pt}
\section{How to Peel with a Knife}
\vspace{-5pt}

We present a two-stage framework to learn highly challenging fine-grained manipulation tasks, exemplified by knife-based peeling. The first stage trains a policy that can maintain blade-object contact across produce variation; the second stage refines the quality of policy's contact behavior using a learned preference-based reward model. This section outlines the three main components of our final system: (1) system design; (2) efficient data collection and policy training to learn a generalizable policy (achieving at least 60\% success rates); and (3) preference-based finetuning that uses a learned human preference reward model to improve the policy.
The result is a system capable of peeling produce of various physical properties with a knife, achieving 100\% success rates on seen produce and over 70\% average success rates on selected held-out produce. Our system and framework are visualized in Figure~\ref{fig:teaser} and~\ref{fig:overview}.

\subsection{System Design}

\paragraphc{Hardware setup} We use a Kinova Gen3 arm which has seven degrees of freedom (DoF). The arm can be torque-controlled which allows for implementation of an impedance controller. We mount an ATI mini45 force-torque sensor between the tool flange and end-effector. The sensor readings are streamed at 500Hz. To stably hold a knife, we custom design an end-effector knife mount (Figure~\ref{fig:teaser}; CAD rendering in Appendix~\ref{app:learning}). We attach two RealSense D405 wrist cameras near the tool.

\paragraphc{Compliant control} We implement an impedance controller to control the Kinova Gen3 arm. The learned policy does not directly command torques; it outputs 10Hz target commands that are tracked by a 500Hz low-level impedance controller on an Intel NUC. This separation assumes that fast contact stabilization is handled by the controller, while the learned policy provides slower geometric and force-aware corrections along the peel. Our implementation is heavily based on open-source projects~\cite{mitchell2025taskjointspacedualarm,wu2024tidybot}. The detailed compliant controller parameters are listed in Appendix~\ref{app:compliance}.

\subsection{Data Collection and Policy Training}

\paragraphc{Infrastructure} We collect high-quality peeling data using human teleoperation. Specifically, we use a 3Dconnexion SpaceMouse to control the 6-dimensional end-effector pose of the Kinova arm. To produce smoother end-effector motion in Cartesian space, we implement a weighted least-squares IK solver (see Appendix~\ref{app:ik} for details). We record trajectories at the same 10Hz rate used by the learned policy, while the controller continues to run at 500Hz. The collected data include robot joint angles, force-torque sensor readings, and RGBD images from the two wrist cameras.

\paragraphc{Data processing} We post-process the data in real time to obtain observations for policy training. We sub-sample the 500Hz force-torque stream by reading its latest value at each 10Hz policy step. We standardize the force-torque readings by subtracting the mean of first 10 samples from all future readings, and segment the RGBD images on knife and object masks separately. The segmentation masks are obtained from running SAM2~\cite{ravi2024sam2} online. We record proprioception as delta end-effector pose in the end-effector frame, as previous work~\cite{chi2024universal} has demonstrated that using a relative end-effector trajectory as imitation target enables generalization to arbitrary base position.

\paragraphc{Training and inference} With the collected dataset, we learn policies that take vision and force as input and predict proprioception (end-effector action commands). For visual inputs, we convert colored images to grayscale, multiply clipped depth values with binary segmented masks (both knife mask and object mask), concatenate the processed RGB and depth features, and apply random crop augmentation. For force inputs, we normalize the readings to $[-1, 1]$. We encode the visual features with ResNet~\cite{he2016deep} and the force features with MLP. We train the policy using Diffusion Policies~\cite{chi2023diffusion} with a simple MLP-based denoiser network. During inference, we send actions at 10Hz.

\subsection{Policy Finetuning with Preference-based Reward}
\label{sec:finetune}

\paragraphc{Reward design} The quality of a peel is difficult to capture with a single objective metric. Furthermore, different observers may weight multiple factors differently -- e.g. peel thickness, continuity and smoothness, and the presence of defects. Among these, peel thickness has a clear geometric definition locally and provides a relatively objective signal, whereas properties such as visual uniformity and overall continuity are more long-horizon and rely strongly on human perceptual judgment.
To capture both aspects, we use a hybrid reward with two annotation resolutions, consisting of a short-horizon quantitative component and a long-horizon qualitative component.
The \uline{quantitative} component measures local relative thickness of the peel. For this, we temporally segment each trajectory at 2Hz, and annotate each segment with one of the six thickness categories shown in Appendix~\ref{app:rewardfigs}.
The \uline{qualitative} component captures holistic human preferences based on the overall visual appearance of the peel. These preferences are difficult to express using local or low-level metrics and are typically global in nature; accordingly, we assign a trajectory-level preference score using a Likert-type ordinal scale, as illustrated in Appendix~\ref{app:rewardfigs}. The qualitative score is not a coarse re-estimate of thickness: it captures long-horizon continuity, smoothness, and defects that the 2Hz thickness labels cannot.
All labels are provided by a single expert annotator applying this fixed rubric.
Both label-to-reward mappings are structured lookup tables ordered by physical desirability, sharing the same structure across produce (Appendix~\ref{app:reward}). We combine the two components using a weighted sum to produce a per-step reward signal that reflects both local geometric precision and global perceptual quality. Implementation details and weighting choices are provided in Appendix~\ref{app:reward}.

\paragraphc{Reward-guided policy finetuning}
We finetune the peeling policy by freezing the base diffusion policy $\pi_{\mathrm{base}}$ and learning a residual policy that predicts action corrections guided by human preference. The frozen components include the base policy and its observation encoder; the learned components in this stage are the reward model and the residual policy.
To enable preference-aware refinement, we first train a reward model offline prior to policy finetuning. The reward model $r_{\psi}(z_t, a_t)$ predicts a human preference score for a state-action pair, where $a_t$ denotes the action recorded in the offline dataset, and $z_t$ denotes the encoded latent feature produced by the frozen base policy observation encoder. It is implemented as a three-layer MLP and trained using a mean squared error objective to regress the normalized reward $r_t$ derived from raw human annotations:
\(\mathcal{L}_{\mathrm{reward}} = \mathbb{E}_{(z_t, a_t)} [\| r_{\psi}(z_t, a_t) - r_t \|^2]\).
In addition to the scalar reward prediction, the reward model exposes an intermediate hidden representation $h_t \in \mathbb{R}^d$, which is later used to condition the residual policy.

The residual policy $\pi_{\mathrm{res}}$, implemented as a two-layer MLP, predicts an action correction conditioned on the base policy's latent feature $z_t$, the base action $a_t^{\mathrm{base}}$, and the reward model's hidden representation $h_t$: $a_t^{\mathrm{res}} = \pi_{\mathrm{res}}(z_t, a_t^{\mathrm{base}}, h_t)$.
The final action executed by the robot is obtained by adding the residual correction to the base action: $a_t^{\mathrm{final}} = a_t^{\mathrm{base}} + a_t^{\mathrm{res}}$.

We train the residual policy using a reward-weighted behavioral cloning objective that encourages the predicted residual to match the difference between the dataset action and the base action:
\begin{equation}
    \mathcal{L}_{\mathrm{res}}
    =
    \mathbb{E}_t \!\left[
        w_t \,
        \big\|
            a_t^{\mathrm{res}} - (a_t - a_t^{\mathrm{base}})
        \big\|^2
    \right]
    + \lambda \,\mathbb{E}_t \!\left[\|a_t^{\mathrm{res}}\|^2\right].
\end{equation}
The first term prioritizes imitating high-quality corrections, while the second term regularizes the magnitude of the residual action to prevent overcorrection. The per-step weight is defined as
$w_t = \exp(\beta r_t)\big/\mathbb{E}_t[\exp(\beta r_t)]$ which emphasizes samples with higher predicted preference.
This objective assumes that high-quality demonstrations contain useful local corrections around the base policy's behavior; the residual parameterization limits the finetuned policy to additive action adjustments rather than re-learning the full action distribution from scratch.

\vspace{-5pt}
\section{Experiments}\vspace{-5pt}
\label{sec:experiments}

We evaluate the proposed framework by answering three questions: whether the base policy can execute knife peeling on real produce, which sensing and data-collection choices are crucial, and whether preference-based finetuning improves the quality of successful executions.

\paragraphc{Task definition} The peeling task requires removing a thin, continuous layer of outer skin from a produce item using a handheld knife, while aligning with human preferences over peel thickness, continuity, smoothness, and efficiency. This task is challenging due to the subtle and highly variable boundary between skin and edible flesh, which demands precise force modulation and stable tool-object contact throughout the motion. A high-quality peel removes a consistent layer of skin with minimal thickness, avoids cutting into the underlying flesh, and maintains smooth, energy-efficient knife motion without jitter or discontinuities. To standardize evaluation across irregularly shaped produce, we define a peel segment as a single stroke executed along the principal axis of the object (i.e. the longest line passing through its centroid). Each peeling trial consists of one or more such strokes, executed sequentially around the circumference until a full side of the surface is covered. While each demonstration contains a single stroke, deployment is autonomous multi-stroke peeling, and each trial is evaluated as one episode.

\paragraphc{Evaluation metrics} We evaluate peeling performance using both qualitative and quantitative metrics derived from human preference and perception, as illustrated in Appendix~\ref{app:rewardfigs}. The \uline{qualitative} metric is a holistic preference score reflecting human judgments of overall peel thickness, length, and continuity. The \uline{quantitative} metric measures local peel thickness as perceived from the robot's onboard sensory data. We consider a peel with a qualitative score greater than 3 to be successful.

We use 10 trials for the data-collection comparison, produce generalization tests, and sample-efficiency study, and 5 trials for the finetuning comparisons and camera and modality ablations. We report success rates with 95\% Wilson score confidence intervals~\cite{wilson1927probable} in brackets; differences with overlapping intervals should be interpreted with caution. Implementation details and hyperparameters are provided in Appendix~\ref{app:training}.

\subsection{Overall Performance}

\paragraphc{Task success rates and generalization}
We evaluate peeling on three produce types: cucumbers, apples, and potatoes. For generalization, we perform two kinds of tests: (1) on \uline{same produce type} as training data, test generalization of the peeling behavior on different start poses and diverse produce instances with small variations in size, shape, stiffness, and surface texture; (2) on \uline{unseen produce types}, test transfer to selected held-out produce with different geometry, compliance, and texture. For all tests with same produce type, we report a success rate of 100\%\ci{72}{100} across cucumber, apple, and potato. Our cucumber policy achieves 50\%\ci{24}{76} success rate on zucchini, apple policy 90\%\ci{60}{98} on pear, and potato policy 80\%\ci{49}{94} on daikon radish. This shows transfer to selected held-out produce rather than general cross-category transfer (see Section~\ref{sec:failure}). Transfer to the visually similar zucchini is the weakest; we hypothesize that its firmer skin and higher pulp compliance shift the contact-force profile more than for pear or daikon. Videos are available on our project website.

\subsection{How to Collect High-Quality Data for Peeling?}
\begin{table}[t]
\centering
\begin{minipage}[c]{0.5\linewidth}
    \centering
    \footnotesize
    \setlength{\tabcolsep}{2.3pt}
    \begin{tabular}{lccc}
    \toprule
    \textbf{Method} & \textbf{Succ.} & \textbf{Score} & \textbf{Time} \\
    \midrule
    Heuristic Planner & 20\%\ci{6}{51}   & 1.8 & 50 \\
    VR                & 20\%\ci{6}{51}   & 2.1 & 69.5 \\
    SpaceMouse \textbf{(ours)} & 100\%\ci{72}{100} & 8.5 & 46 \\
    \midrule
    Kinesthetic Teaching & 100\%\ci{72}{100} & 7.2 & 13.8 \\
    Replay               & 0\%\ci{0}{28}     & 1.6 & N.A. \\
    \bottomrule
    \end{tabular}
\end{minipage}%
\hfill
\begin{minipage}[c]{0.46\linewidth}
    \caption{\small \textbf{Data-collection quality and efficiency.} Each score is averaged over 10 cucumber trajectories; brackets show 95\% confidence intervals. Only trajectories with score above 3 are counted as successes. Average time taken is only calculated from successful trajectories.}
    \label{table:data}
\end{minipage}
\end{table}

\paragraphc{Comparison of data collection methods} We compare our SpaceMouse-based teleoperation method with model-based planner, VR-based teleoperation~\cite{lin2024learning}, and kinesthetic teaching followed by replay with heuristic-based compliance parameters~\cite{hou2025adaptive}. For each data collection method, we evaluate the quality of 10 trajectories collected using qualitative metrics defined in Section~\ref{sec:finetune}. We show the success rate, average performance and time taken for each method in Table~\ref{table:data}; it should be interpreted as a comparison of data collection pipelines under our hardware and controller rather than as a general ranking of teleoperation interfaces.
The {model-based planner} plans a knife pose trajectory along the object surface from a third-view point cloud (details in Appendix~\ref{app:datacollection}).
Since this baseline explicitly encodes a surface-following heuristic, it can peel with human-in-the-loop correction of the next desired knife pose. However, without online correction, it easily fails under variations of object geometries and surface properties -- motivating a closed-loop, fully autonomous policy.
Details on other data collection methods are provided in Appendix~\ref{app:datacollection}.
In our setup, SpaceMouse teleoperation -- without additional modifications like adaptive compliance or haptic feedback -- provides an efficient way to collect high-quality data for the challenging task of peeling with a knife.

\subsection{How to Learn High-Performance Peeling Policies?}

Ablation studies on camera placement (Table~\ref{tab:wrist_cam}), data modalities (Table~\ref{tab:modalities}), and sample efficiency (Table~\ref{tab:efficiency}) are detailed in Appendix~\ref{app:learning}; grayscale vision and force-torque sensing each improve both success rate and average score, and using both wrist cameras yields the highest average score.

\subsection{How to Align Learned Policies with Human Preference?}
\begin{table}[t]
\centering
\footnotesize
\setlength{\tabcolsep}{2pt}
\begin{minipage}[t]{0.48\linewidth}
    \centering
    \begin{tabular}{lcccc}
    \toprule
    & \multicolumn{2}{c}{\textbf{Apple}} & \multicolumn{2}{c}{\textbf{Potato}} \\
    \cmidrule(lr){2-3}\cmidrule(lr){4-5}
    \textbf{Method} & \textbf{Succ.} & \textbf{Score} & \textbf{Succ.} & \textbf{Score} \\
    \midrule
    \textbf{Ours} & 100\ci{57}{100} & 7.1 & 100\ci{57}{100} & 7.3 \\
    Base Only     & 60\ci{23}{88}   & 3.8 & 80\ci{38}{96}   & 5.9 \\
    Quan. R.      & 40\ci{12}{77}   & 3.8 & 60\ci{23}{88}   & 4.0 \\
    Qual. R.      & 0\ci{0}{43}     & 1.4 & 60\ci{23}{88}   & 4.4 \\
    IQL-w         & 0\ci{0}{43}     & 0.2 & 0\ci{0}{43}     & 2.6 \\
    \bottomrule
    \end{tabular}
    \caption{\small \textbf{Finetuning baseline comparison.}
    Success rates (\%) with 95\% confidence intervals and average scores on apples and potatoes, over 5 trials each.}
    \label{table:baseline}
\end{minipage}%
\hfill
\begin{minipage}[t]{0.48\linewidth}
    \centering
    \begin{tabular}{lcccc}
    \toprule
    & \multicolumn{2}{c}{\textbf{Apple}} & \multicolumn{2}{c}{\textbf{Potato}} \\
    \cmidrule(lr){2-3}\cmidrule(lr){4-5}
    \textbf{Reward} & \textbf{Succ.} & \textbf{Score} & \textbf{Succ.} & \textbf{Score} \\
    \midrule
    \textbf{Ours} & 100\ci{57}{100} & 7.1 & 100\ci{57}{100} & 7.3 \\
    PS+B          & 0\ci{0}{43}     & 0.0 & 0\ci{0}{43}     & 0.0 \\
    P60+FG        & 0\ci{0}{43}     & 1.0 & 20\ci{4}{62}    & 3.8 \\
    PT+FG         & 40\ci{12}{77}   & 2.3 & 0\ci{0}{43}     & 1.2 \\
    \bottomrule
    \end{tabular}
    \caption{\small\textbf{Reward design.} Reward density over time -- per-step (PS), per-60-steps (P60), or per-trajectory (PT) -- and in value -- binary (B) or fine-grained (FG). Ours is PS+FG.}
    \label{table:reward_design}
\end{minipage}

\vspace{0.8em}

\begin{minipage}[t]{0.48\linewidth}
    \centering
    \begin{tabular}{lcccc}
    \toprule
    & \multicolumn{2}{c}{\textbf{Apple}} & \multicolumn{2}{c}{\textbf{Potato}} \\
    \cmidrule(lr){2-3}\cmidrule(lr){4-5}
    \textbf{Method} & \textbf{Succ.} & \textbf{Score} & \textbf{Succ.} & \textbf{Score} \\
    \midrule
    \textbf{Ours} & 100\ci{57}{100} & 7.1 & 100\ci{57}{100} & 7.3 \\
    One-Step      & 0\ci{0}{43}     & 1.2 & 60\ci{23}{88}   & 4.6 \\
    Filter        & 0\ci{0}{43}     & 1.0 & 80\ci{38}{96}   & 5.2 \\
    \bottomrule
    \end{tabular}
    \caption{\small\textbf{Using the learned reward.} One-step advantage instead of raw reward (One-Step), and finetuning only on high-reward steps selected by a hard threshold (Filter).}
    \label{table:ft_design}
\end{minipage}%
\hfill
\begin{minipage}[t]{0.48\linewidth}
    \centering
    \begin{tabular}{lcccc}
    \toprule
    & \multicolumn{2}{c}{\textbf{Apple}} & \multicolumn{2}{c}{\textbf{Potato}} \\
    \cmidrule(lr){2-3}\cmidrule(lr){4-5}
    \textbf{Training} & \textbf{Succ.} & \textbf{Score} & \textbf{Succ.} & \textbf{Score} \\
    \midrule
    \textbf{Ours} & 100\ci{57}{100} & 7.1 & 100\ci{57}{100} & 7.3 \\
    Scratch       & 0\ci{0}{43}     & 0.4 & 0\ci{0}{43}     & 0.0 \\
    No Res.       & 0\ci{0}{43}     & 1.2 & 40\ci{12}{77}   & 3.0 \\
    \bottomrule
    \end{tabular}
    \caption{\small\textbf{Training scheme.} Reward-weighted training from scratch (Scratch) and finetuning the base policy without a residual network (No Res.), vs.\ our frozen base policy with a residual network.}
    \label{table:residual}
\end{minipage}
\end{table}

\paragraphc{Reward model validation} On a held-out test set (20\% of trajectories), the learned reward model closely tracks the human labels (Spearman $\rho = 0.85$, Pearson $r = 0.88$, over 90\% pairwise ranking accuracy). Samples where it deviates from the human labels are mostly ones the annotator also found ambiguous; we attribute this robustness to the labels reflecting a holistic assessment of concrete factors (e.g. peel thickness and length) rather than arbitrary taste.

\begin{wrapfigure}{r}{0.42\linewidth}
    \centering
    \includegraphics[width=\linewidth]{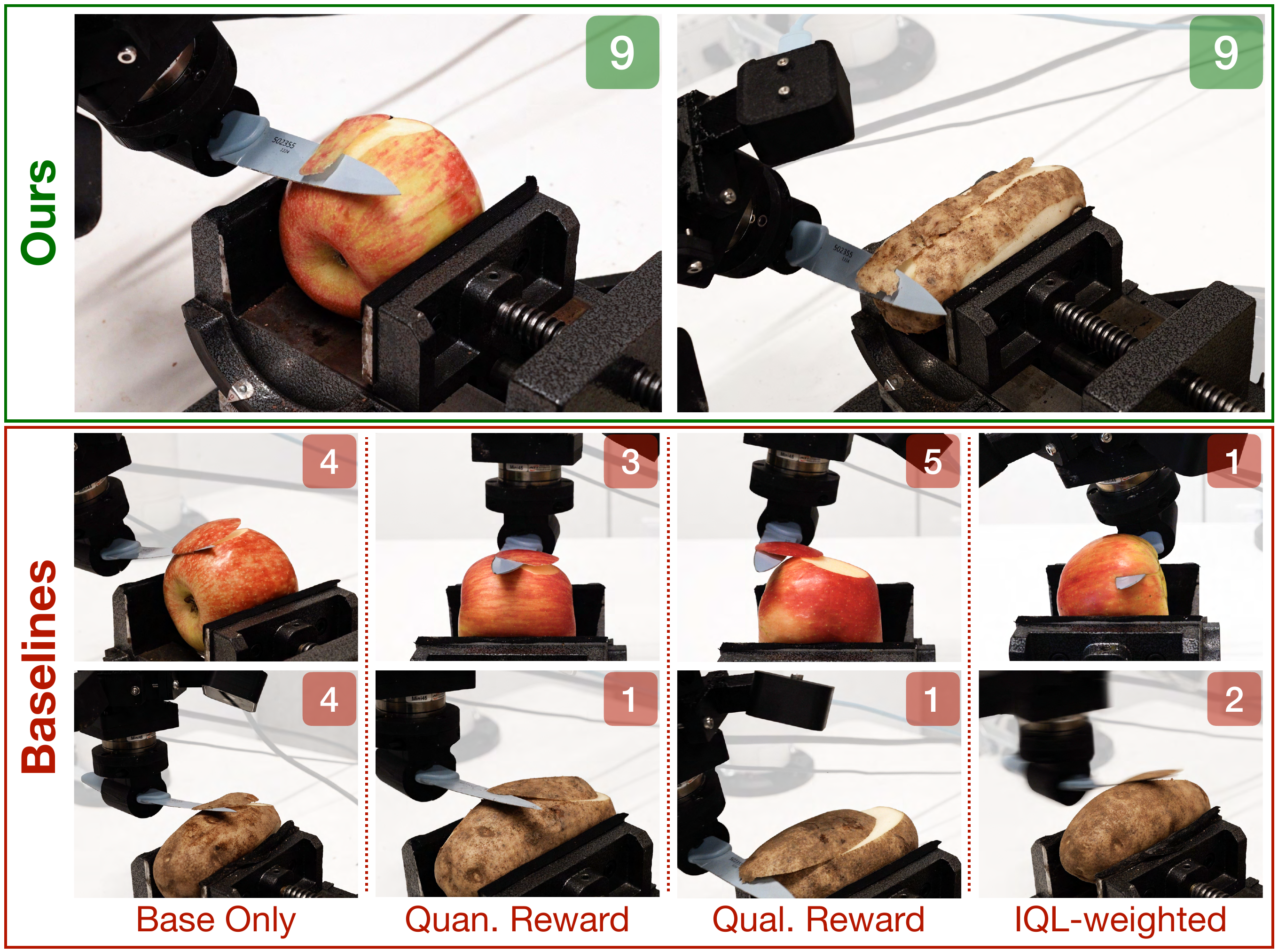}
    \caption{\small\textbf{Qualitative comparison with baselines.}
    The peel quality of our policy is substantially higher than that of the baselines, even when both are counted as success,
    as indicated by the qualitative score (top right corner of each image).}
    \label{fig:baseline}
\end{wrapfigure}

\paragraphc{Comparison with baselines} Our final policy is finetuned with both quantitative and qualitative rewards. To ensure fairness of comparison, we train all policies with the same dataset and compare against four baselines to isolate the role of finetuning and training objective: \uline{Base Only}, the base policy without finetuning; \uline{Quantitative Reward}, which finetunes using only the quantitative preference reward; \uline{Qualitative Reward}, which finetunes using only the qualitative preference reward; and \uline{IQL-weighted}, which trains critic networks following IQL~\cite{kostrikov2021offline}, estimates $Q$ and $V$, computes per-step advantages as $A=Q-V$, and uses advantage-weighted supervised finetuning to emphasize actions that are relatively better than the policy's expected behavior at the same state.
We choose IQL-weighted as the external baseline because it uses the same learned reward model as our approach: our goal is to isolate the effect of the two-stage residual formulation under a shared reward rather than to propose a new preference learning algorithm, which we complement with ablations (Tables~\ref{table:reward_design}--\ref{table:residual}).
We show quantitative results in Table~\ref{table:baseline} and qualitative comparison in Figure~\ref{fig:baseline}.
In these experiments, reward-weighted supervised finetuning gives better performance than the offline RL-style critic-guided finetuning variant with implicit Q-learning~\cite{kostrikov2021offline}, while exhibiting better training stability. Using both reward components is also necessary: on apple, finetuning with only the quantitative or only the qualitative reward reaches 40\% and 0\% success, respectively.
Results on apple are close to binary across Tables~\ref{table:baseline}--\ref{table:residual}, unlike potato: apple peels have a narrower range of acceptable thickness, a more slippery skin, and larger curvature, so suboptimal policies fail immediately, whereas potato peels tolerate more imprecision.

\paragraphc{Designing human preference scores} We explore two main design choices for representation of human preference: density across time horizon (per-step vs per-60-steps vs per-trajectory) and density in value (binary vs fine-grained). While denser rewards might provide better local credit assignment, they increase annotation effort. Empirically, we find the right balance lies at per-step, fine-grained reward values -- where labor cost can be offset by learning a reward function from a small number of hand-annotated data (see details in Section~\ref{sec:finetune}). In Table~\ref{table:reward_design}, we show the quantitative comparison of finetuned policy performance from different reward designs.

\paragraphc{Utilizing learned reward model} Given a reward model that can assign a per-step score to a trajectory, a key question is how to convert this learning signal into a policy update effectively. In addition to our final approach, we compare two alternatives: (A) \uline{One-Step Advantage}: instead of using the raw reward, we calculate a trajectory-centered ``advantage'' by first computing a trajectory baseline $b_\tau$, then forming a per-step advantage $A_t = r_t - b_\tau$, enabling more localized credit assignment when weighting behavior cloning updates. (B) \uline{Binary Filtering}: instead of applying smooth per-step weighting via an exponential preference-based weight, we hard-select high-scoring steps using a binary filter, retaining only a fixed top fraction according to the preference reward, and finetune uniformly on the remaining samples without any additional weighting. Empirical results are shown in Table~\ref{table:ft_design}; our approach performs best among these tested ways of using the learned reward.

\paragraphc{Choosing the right training scheme} Finally, we investigate two important design choices regarding training: (1) whether to use a residual network, and (2) whether to finetune on a frozen base policy or train from scratch. Our final policy freezes the base policy and learns only the residual correction. We compare the performance with two baselines: (1) \uline{No Residual}: directly finetuning the base policy; (2) \uline{From Scratch}: learning the policy from scratch with reward weighting. Empirical results are shown in Table~\ref{table:residual}. Both the two-stage training and the residual network are important for stable performance. We attribute this to the residual parameterization preserving the contact behavior of the frozen base policy, whereas the alternatives must re-learn the full action distribution from limited data.

\subsection{Failure Cases}
\label{sec:failure}

We systematically collect and study the failure cases. In addition to cutting too low and cutting too high (qualitative scores 1 and 2), failures occur when the learned policy encounters produce whose geometry, compliance, or surface texture changes the contact condition beyond what is represented in the demonstrations. For example, in our video, we show failures when deploying cucumber policy to apple, apple to potato, and potato to cucumber. These failures indicate that the method still assumes overlap between the training demonstrations and deployment conditions; extending transfer across larger material and shape shifts remains open.

\vspace{-5pt}
\section{Conclusion}
\vspace{-5pt}

In this work, we propose a two-stage approach to learn knife-based peeling from real-world demonstrations and human preference labels. The key idea is to first learn a contact-maintaining peeling skill from force-aware demonstration data, then refine the policy's motion quality with a reward model learned from human-evaluated quality labels. Across cucumbers, apples, and potatoes, our experiments show same-produce success from 50--200 trajectories, zero-shot transfer to selected held-out produce, and improved apple and potato performance after preference-based finetuning.

\vspace{-5pt}
\section{Limitations and Future Work}\vspace{-5pt}
\label{sec:future}

While our framework demonstrates promising performance on knife-based peeling, a key limitation is its reliance on manually collected, high-quality demonstrations. Improving scalability is therefore an important direction for future work. One promising extension is to incorporate online reinforcement learning as a finetuning stage, which recent work has shown to be effective for refining real-world manipulation policies~\cite{ankile2025residual,xiao2025self,li2025gr}. Another direction is to reduce reliance on fully human teleoperation by adopting mixed-autonomy data collection, for example by combining model-based planners with intermittent human intervention, thereby lowering human effort while maintaining data quality. The reward design also involves manual effort: the label-to-reward mappings and weights are set per produce type, and all labels come from a single expert annotator; adapting rewards to new produce or other users, and modeling annotator disagreement, remain open. In addition, our comparisons use 5 or 10 real-world trials per condition, limiting confidence in smaller performance differences.

Beyond scalability, several modifications could further improve overall system performance. First, inspired by advances in large language model alignment, more expressive reward parameterizations -- such as ranking-based or listwise rewards -- may enable stronger alignment with human notions of task quality. Second, perception remains a bottleneck: augmenting the sensing setup with additional viewpoints (e.g. a front-facing camera) could improve estimation of peel thickness and surface quality. While orthogonal to this work, continued progress in depth sensing hardware and segmentation models would further benefit such systems.

More broadly, by introducing both qualitative and quantitative evaluation metrics for knife-based peeling and demonstrating how to learn an effective reward model from them, this work opens the door to systematic studies of the trade-off between data quality and data quantity in preference-based robot learning.

Finally, a practical limitation of real-world food manipulation research is the generation of food waste. We hope future work will explore reusable ``surrogate produce'' or improved simulation and sim-to-real transfer methods that enable similar experimentation with reduced environmental cost.

% Acknowledgments are suppressed in the anonymous submission version (corl_2026.sty
% only emits this block when compiled with [final] or [preprint]).
\acknowledgments{We thank the authors of the open-source repositories that informed our implementation of the compliant controller on the Kinova Gen3~\cite{wu2024tidybot,madan_gen3_2024,mitchell2025taskjointspacedualarm} for their technical guidance. We are grateful to Yifan Hou for initial advice on compliant controller implementation and for sharing resources related to mount design. We also thank Pingchuan Ma for assistance with photography and videography.}

\bibliography{references}

@inproceedings{chi2023diffusion,
  title={Diffusion policy: Visuomotor policy learning via action diffusion},
  author={Chi, Cheng and Feng, Siyuan and Du, Yilun and Xu, Zhenjia and Cousineau, Eric and Burchfiel, Benjamin and Song, Shuran},
  booktitle={RSS},
  year={2023}
}

@inproceedings{lin2024learning,
  author={Lin, Toru and Zhang, Yu and Li, Qiyang and Qi, Haozhi and Yi, Brent and Levine, Sergey and Malik, Jitendra},
  title={Learning Visuotactile Skills with Two Multifingered Hands},
  booktitle={2025 IEEE International Conference on Robotics and Automation (ICRA)},
  pages={5637--5643},
  year={2025},
  organization={IEEE}
}

@inproceedings{he2016deep,
  title={Deep residual learning for image recognition},
  author={He, Kaiming and Zhang, Xiangyu and Ren, Shaoqing and Sun, Jian},
  booktitle={CVPR},
  year={2016}
}

@inproceedings{xue2025reactive,
  title={Reactive diffusion policy: Slow-fast visual-tactile policy learning for contact-rich manipulation},
  author={Xue, Han and Ren, Jieji and Chen, Wendi and Zhang, Gu and Fang, Yuan and Gu, Guoying and Xu, Huazhe and Lu, Cewu},
  booktitle={Proceedings of Robotics: Science and Systems (RSS)},
  year={2025}
}

@article{chen2024vegetable,
  title={Vegetable peeling: A case study in constrained dexterous manipulation},
  author={Chen, Tao and Cousineau, Eric and Kuppuswamy, Naveen and Agrawal, Pulkit},
  journal={arXiv preprint arXiv:2407.07884},
  year={2024}
}

@misc{mitchell2025taskjointspacedualarm,
      title={Task and Joint Space Dual-Arm Compliant Control}, 
      author={Alexander L. Mitchell and Tobit Flatscher and Ingmar Posner},
      year={2025},
      eprint={2504.21159},
      archivePrefix={arXiv},
      primaryClass={cs.RO},
      url={https://arxiv.org/abs/2504.21159}, 
}

@misc{madan_gen3_2024,
  author       = {Madan, Rishabh and Jenamani, Rajat Kumar and Han, Seo Wook},
  title        = {{gen3\_compliant\_controllers}: {ROS} Package providing compliant controllers for the {Kinova Gen3} arm},
  year         = {2024},
  publisher    = {GitHub},
  journal      = {GitHub repository},
  howpublished = {\url{https://github.com/empriselab/gen3_compliant_controllers}},
  commit       = {main}
}

@inproceedings{ravi2024sam2,
  title={{SAM} 2: Segment Anything in Images and Videos},
  author={Ravi, Nikhila and Gabeur, Valentin and Hu, Yuan-Ting and Hu, Ronghang and Ryali, Chaitanya and Ma, Tengyu and Khedr, Haitham and R{\"a}dle, Roman and Rolland, Chloe and Gustafson, Laura and Mintun, Eric and Pan, Junting and Alwala, Kalyan Vasudev and Carion, Nicolas and Wu, Chao-Yuan and Girshick, Ross and Doll{\'a}r, Piotr and Feichtenhofer, Christoph},
  booktitle={International Conference on Learning Representations},
  pages={28085--28128},
  year={2025}
}

@inproceedings{lin2025sim,
  title={Sim-to-real reinforcement learning for vision-based dexterous manipulation on humanoids},
  author={Lin, Toru and Sachdev, Kartik and Fan, Linxi and Malik, Jitendra and Zhu, Yuke},
  booktitle={Proceedings of The 9th Conference on Robot Learning},
  series={Proceedings of Machine Learning Research},
  volume={305},
  pages={4926--4940},
  year={2025},
  publisher={PMLR}
}

@article{fang2023anygrasp,
  title={{AnyGrasp}: Robust and efficient grasp perception in spatial and temporal domains},
  author={Fang, Hao-Shu and Wang, Chenxi and Fang, Hongjie and Gou, Minghao and Liu, Jirong and Yan, Hengxu and Liu, Wenhai and Xie, Yichen and Lu, Cewu},
  journal={IEEE Transactions on Robotics},
  volume={39},
  number={5},
  pages={3929--3945},
  year={2023},
  publisher={IEEE}
}

@inproceedings{wang2022dexgraspnet,
  title={{DexGraspNet}: A large-scale robotic dexterous grasp dataset for general objects based on simulation},
  author={Wang, Ruicheng and Zhang, Jialiang and Chen, Jiayi and Xu, Yinzhen and Li, Puhao and Liu, Tengyu and Wang, He},
  booktitle={2023 IEEE International Conference on Robotics and Automation (ICRA)},
  pages={11359--11366},
  year={2023},
  organization={IEEE}
}

@article{intelligence2025pi,
  title={$\pi_{0.5}$: a Vision-Language-Action Model with Open-World Generalization},
  author={{Physical Intelligence} and Black, Kevin and Brown, Noah and Darpinian, James and Dhabalia, Karan and Driess, Danny and Esmail, Adnan and Equi, Michael and Finn, Chelsea and Fusai, Niccolo and others},
  journal={arXiv preprint arXiv:2504.16054},
  year={2025}
}

@inproceedings{lum2024dextrah,
  title={{DextrAH-G}: Pixels-to-action dexterous arm-hand grasping with geometric fabrics},
  author={Lum, Tyler Ga Wei and Matak, Martin and Makoviychuk, Viktor and Handa, Ankur and Allshire, Arthur and Hermans, Tucker and Ratliff, Nathan D and Van Wyk, Karl},
  booktitle={Proceedings of The 8th Conference on Robot Learning},
  series={Proceedings of Machine Learning Research},
  volume={270},
  pages={3182--3211},
  year={2025},
  publisher={PMLR}
}

@inproceedings{chi2024universal,
title={Universal Manipulation Interface: In-The-Wild Robot Teaching Without In-The-Wild Robots},
author={Chi, Cheng and Xu, Zhenjia and Pan, Chuer and Cousineau, Eric and Burchfiel, Benjamin and Feng, Siyuan and Tedrake, Russ and Song, Shuran},
booktitle={Proceedings of Robotics: Science and Systems (RSS)},
year={2024}
}

@inproceedings{hou2025adaptive,
  title={Adaptive compliance policy: Learning approximate compliance for diffusion guided control},
  author={Hou, Yifan and Liu, Zeyi and Chi, Cheng and Cousineau, Eric and Kuppuswamy, Naveen and Feng, Siyuan and Burchfiel, Benjamin and Song, Shuran},
  booktitle={2025 IEEE International Conference on Robotics and Automation (ICRA)},
  pages={4829--4836},
  year={2025},
  organization={IEEE}
}

@article{chen2025dexforce,
  title={{DexForce}: Extracting force-informed actions from kinesthetic demonstrations for dexterous manipulation},
  author={Chen, Claire and Yu, Zhongchun and Choi, Hojung and Cutkosky, Mark and Bohg, Jeannette},
  journal={IEEE Robotics and Automation Letters},
  year={2025},
  publisher={IEEE}
}

@inproceedings{ye2024morpheus,
  title={{MORPHeus}: a Multimodal One-armed Robot-assisted Peeling system with Human Users in-the-loop},
  author={Ye, Ruolin and Hu, Yifei and Bian, Yuhan Anjelica and Kulm, Luke and Bhattacharjee, Tapomayukh},
  booktitle={2024 IEEE International Conference on Robotics and Automation (ICRA)},
  pages={9540--9547},
  year={2024},
  organization={IEEE}
}

@inproceedings{straivzys2020surfing,
  title={Surfing on an uncertain edge: Precision cutting of soft tissue using torque-based medium classification},
  author={Strai{\v{z}}ys, Art{\=u}ras and Burke, Michael and Ramamoorthy, Subramanian},
  booktitle={2020 IEEE International Conference on Robotics and Automation (ICRA)},
  pages={4623--4629},
  year={2020},
  organization={IEEE}
}

@article{ankile2025residual,
  title={Residual Off-Policy {RL} for Finetuning Behavior Cloning Policies},
  author={Ankile, Lars and Jiang, Zhenyu and Duan, Rocky and Shi, Guanya and Abbeel, Pieter and Nagabandi, Anusha},
  journal={arXiv preprint arXiv:2509.19301},
  year={2025}
}

@article{li2025gr,
  title={{GR-RL}: Going Dexterous and Precise for Long-Horizon Robotic Manipulation},
  author={Li, Yunfei and Ma, Xiao and Xu, Jiafeng and Cui, Yu and Cui, Zhongren and Han, Zhigang and Huang, Liqun and Kong, Tao and Liu, Yuxiao and Niu, Hao and others},
  journal={arXiv preprint arXiv:2512.01801},
  year={2025}
}

@article{xiao2025self,
  title={Self-Improving Vision-Language-Action Models with Data Generation via Residual {RL}},
  author={Xiao, Wenli and Lin, Haotian and Peng, Andy and Xue, Haoru and He, Tairan and Xie, Yuqi and Hu, Fengyuan and Wu, Jimmy and Luo, Zhengyi and Fan, Linxi and others},
  journal={arXiv preprint arXiv:2511.00091},
  year={2025}
}

@inproceedings{kostrikov2021offline,
  title={Offline reinforcement learning with implicit {Q}-learning},
  author={Kostrikov, Ilya and Nair, Ashvin and Levine, Sergey},
  booktitle={International Conference on Learning Representations},
  year={2022}
}

@article{watanabe2013cooking,
  title={Cooking behavior with handling general cooking tools based on a system integration for a life-sized humanoid robot},
  author={Watanabe, Yoshiaki and Nagahama, Kotaro and Yamazaki, Kimitoshi and Okada, Kei and Inaba, Masayuki},
  journal={Paladyn, Journal of Behavioral Robotics},
  volume={4},
  number={2},
  pages={63--72},
  year={2013},
  publisher={Versita}
}

@inproceedings{dong2021food,
  title={Food peeling method for dual-arm cooking robot},
  author={Dong, Chenyu and Yu, Liangliang and Takizawa, Masaru and Kudoh, Shunsuke and Suehiro, Takashi},
  booktitle={2021 IEEE/SICE International Symposium on System Integration (SII)},
  pages={801--806},
  year={2021},
  organization={IEEE}
}

@inproceedings{liu2025forcemimic,
  title={{ForceMimic}: Force-centric imitation learning with force-motion capture system for contact-rich manipulation},
  author={Liu, Wenhai and Wang, Junbo and Wang, Yiming and Wang, Weiming and Lu, Cewu},
  booktitle={2025 IEEE International Conference on Robotics and Automation (ICRA)},
  pages={1105--1112},
  year={2025},
  organization={IEEE}
}

@article{he2025foar,
  title={{FoAR}: Force-Aware Reactive Policy for Contact-Rich Robotic Manipulation},
  author={He, Zihao and Fang, Hongjie and Chen, Jingjing and Fang, Hao-Shu and Lu, Cewu},
  journal={IEEE Robotics and Automation Letters},
  year={2025},
  publisher={IEEE}
}

@inproceedings{zhang2023efficient,
  title={Efficient sim-to-real transfer of contact-rich manipulation skills with online admittance residual learning},
  author={Zhang, Xiang and Wang, Changhao and Sun, Lingfeng and Wu, Zheng and Zhu, Xinghao and Tomizuka, Masayoshi},
  booktitle={Conference on Robot Learning},
  pages={1621--1639},
  year={2023},
  organization={PMLR}
}

@inproceedings{zhang2024bridging,
  title={Bridging the sim-to-real gap with dynamic compliance tuning for industrial insertion},
  author={Zhang, Xiang and Tomizuka, Masayoshi and Li, Hui},
  booktitle={2024 IEEE International Conference on Robotics and Automation (ICRA)},
  pages={4356--4363},
  year={2024},
  organization={IEEE}
}

@inproceedings{wu2024tidybot,
  title = {{TidyBot++}: An Open-Source Holonomic Mobile Manipulator for Robot Learning},
  author = {Wu, Jimmy and Chong, William and Holmberg, Robert and Prasad, Aaditya and Gao, Yihuai and Khatib, Oussama and Song, Shuran and Rusinkiewicz, Szymon and Bohg, Jeannette},
  booktitle = {Conference on Robot Learning},
  year = {2024}
}

@inproceedings{xu2023roboninja,
	title={{RoboNinja}: Learning an Adaptive Cutting Policy for Multi-Material Objects},
	author={Xu, Zhenjia and Xian, Zhou and Lin, Xingyu and Chi, Cheng and Huang, Zhiao and Gan, Chuang and Song, Shuran},
	booktitle={Proceedings of Robotics: Science and Systems (RSS)},
	year={2023}
}

@article{sochacki2024towards,
  title={Towards practical robotic chef: Review of relevant work and future challenges},
  author={Sochacki, Grzegorz and Zhang, Xiaoping and Abdulali, Arsen and Iida, Fumiya},
  journal={Journal of Field Robotics},
  volume={41},
  number={5},
  pages={1596--1616},
  year={2024},
  publisher={Wiley Online Library}
}

@inproceedings{long2014force,
  title={Force/vision control for robotic cutting of soft materials},
  author={Long, Philip and Khalil, Wisama and Martinet, Philippe},
  booktitle={2014 IEEE/RSJ international conference on intelligent robots and systems},
  pages={4716--4721},
  year={2014},
  organization={IEEE}
}

@article{mu2023dexterous,
  title={Dexterous robotic cutting based on fracture mechanics and force control},
  author={Mu, Xiaoqian and Xue, Yuechuan and Jia, Yan-Bin},
  journal={IEEE Transactions on Automation Science and Engineering},
  year={2023},
  publisher={IEEE}
}

@inproceedings{mu2019robotic,
  title={Robotic cutting: Mechanics and control of knife motion},
  author={Mu, Xiaoqian and Xue, Yuechuan and Jia, Yan-Bin},
  booktitle={2019 International Conference on Robotics and Automation (ICRA)},
  pages={3066--3072},
  year={2019},
  organization={IEEE}
}

@article{jamdagni2024robotic,
  title={Robotic Cutting of Fruits and Vegetables: Modeling the Effects of Deformation, Fracture Toughness, Knife Edge Geometry, and Motion},
  author={Jamdagni, Prajjwal and Jia, Yan-Bin},
  journal={IEEE Transactions on Robotics},
  year={2024},
  publisher={IEEE}
}

@inproceedings{yang2016vision,
  title={Vision-based cutting control of deformable objects},
  author={Yang, Bohan and Wang, Hesheng and Chen, Weidong and Wang, Zehui},
  booktitle={2016 IEEE International Conference on Real-time Computing and Robotics (RCAR)},
  pages={650--655},
  year={2016},
  organization={IEEE}
}

@article{han2020vision,
  title={Vision-based cutting control of deformable objects with surface tracking},
  author={Han, Lijun and Wang, Hesheng and Liu, Zhe and Chen, Weidong and Zhang, Xiufeng},
  journal={IEEE/ASME Transactions on Mechatronics},
  volume={26},
  number={4},
  pages={2016--2026},
  year={2020},
  publisher={IEEE}
}

@article{yin2021modeling,
  title={Modeling, learning, perception, and control methods for deformable object manipulation},
  author={Yin, Hang and Varava, Anastasia and Kragic, Danica},
  journal={Science Robotics},
  volume={6},
  number={54},
  pages={eabd8803},
  year={2021},
  publisher={American Association for the Advancement of Science}
}

@INPROCEEDINGS{heiden2021disect, 
    AUTHOR    = {Eric Heiden AND Miles Macklin AND Yashraj S Narang AND Dieter Fox AND Animesh Garg AND Fabio Ramos}, 
    TITLE     = {{DiSECt: A Differentiable Simulation Engine for Autonomous Robotic Cutting}}, 
    BOOKTITLE = {Proceedings of Robotics: Science and Systems}, 
    YEAR      = {2021}, 
    ADDRESS   = {Virtual}, 
    MONTH     = {July}, 
    DOI       = {10.15607/RSS.2021.XVII.067} 
}

@article{wirth2017survey,
  title={A survey of preference-based reinforcement learning methods},
  author={Wirth, Christian and Akrour, Riad and Neumann, Gerhard and F{\"u}rnkranz, Johannes},
  journal={Journal of Machine Learning Research},
  volume={18},
  number={136},
  pages={1--46},
  year={2017}
}

@article{christiano2017deep,
  title={Deep reinforcement learning from human preferences},
  author={Christiano, Paul F and Leike, Jan and Brown, Tom and Martic, Miljan and Legg, Shane and Amodei, Dario},
  journal={Advances in neural information processing systems},
  volume={30},
  year={2017}
}

@inproceedings{furnkranz2003pairwise,
  title={Pairwise preference learning and ranking},
  author={F{\"u}rnkranz, Johannes and H{\"u}llermeier, Eyke},
  booktitle={European conference on machine learning},
  pages={145--156},
  year={2003},
  organization={Springer}
}

@article{chang2024survey,
  title={A survey on evaluation of large language models},
  author={Chang, Yupeng and Wang, Xu and Wang, Jindong and Wu, Yuan and Yang, Linyi and Zhu, Kaijie and Chen, Hao and Yi, Xiaoyuan and Wang, Cunxiang and Wang, Yidong and others},
  journal={ACM transactions on intelligent systems and technology},
  volume={15},
  number={3},
  pages={1--45},
  year={2024},
  publisher={ACM New York, NY}
}

@inproceedings{hejna2023few,
  title={Few-shot preference learning for human-in-the-loop {RL}},
  author={Hejna III, Donald Joseph and Sadigh, Dorsa},
  booktitle={Conference on Robot Learning},
  pages={2014--2025},
  year={2023},
  organization={PMLR}
}

@inproceedings{lee2021pebble,
  title={{PEBBLE}: Feedback-efficient interactive reinforcement learning via relabeling experience and unsupervised pre-training},
  author={Lee, Kimin and Smith, Laura and Abbeel, Pieter},
  booktitle={Proceedings of the 38th International Conference on Machine Learning},
  series={Proceedings of Machine Learning Research},
  volume={139},
  pages={6152--6163},
  year={2021},
  publisher={PMLR}
}

@inproceedings{hejna2023contrastive,
  title={Contrastive preference learning: learning from human feedback without {RL}},
  author={Hejna, Joey and Rafailov, Rafael and Sikchi, Harshit and Finn, Chelsea and Niekum, Scott and Knox, W Bradley and Sadigh, Dorsa},
  booktitle={International Conference on Learning Representations},
  pages={18770--18798},
  year={2024}
}

@article{ibarz2018reward,
  title={Reward learning from human preferences and demonstrations in {A}tari},
  author={Ibarz, Borja and Leike, Jan and Pohlen, Tobias and Irving, Geoffrey and Legg, Shane and Amodei, Dario},
  journal={Advances in neural information processing systems},
  volume={31},
  year={2018}
}

@article{chen2025fdpp,
  title={{FDPP}: Fine-tune diffusion policy with human preference},
  author={Chen, Yuxin and Jha, Devesh K and Tomizuka, Masayoshi and Romeres, Diego},
  journal={arXiv preprint arXiv:2501.08259},
  year={2025}
}

@article{wilson1927probable,
  title={Probable inference, the law of succession, and statistical inference},
  author={Wilson, Edwin B.},
  journal={Journal of the American Statistical Association},
  volume={22},
  number={158},
  pages={209--212},
  year={1927}
}

\clearpage

\appendix

\section{Reward Metric Visualizations}
\label{app:rewardfigs}

\begin{figure}[h!]
    \centering
    \includegraphics[width=0.55\linewidth]{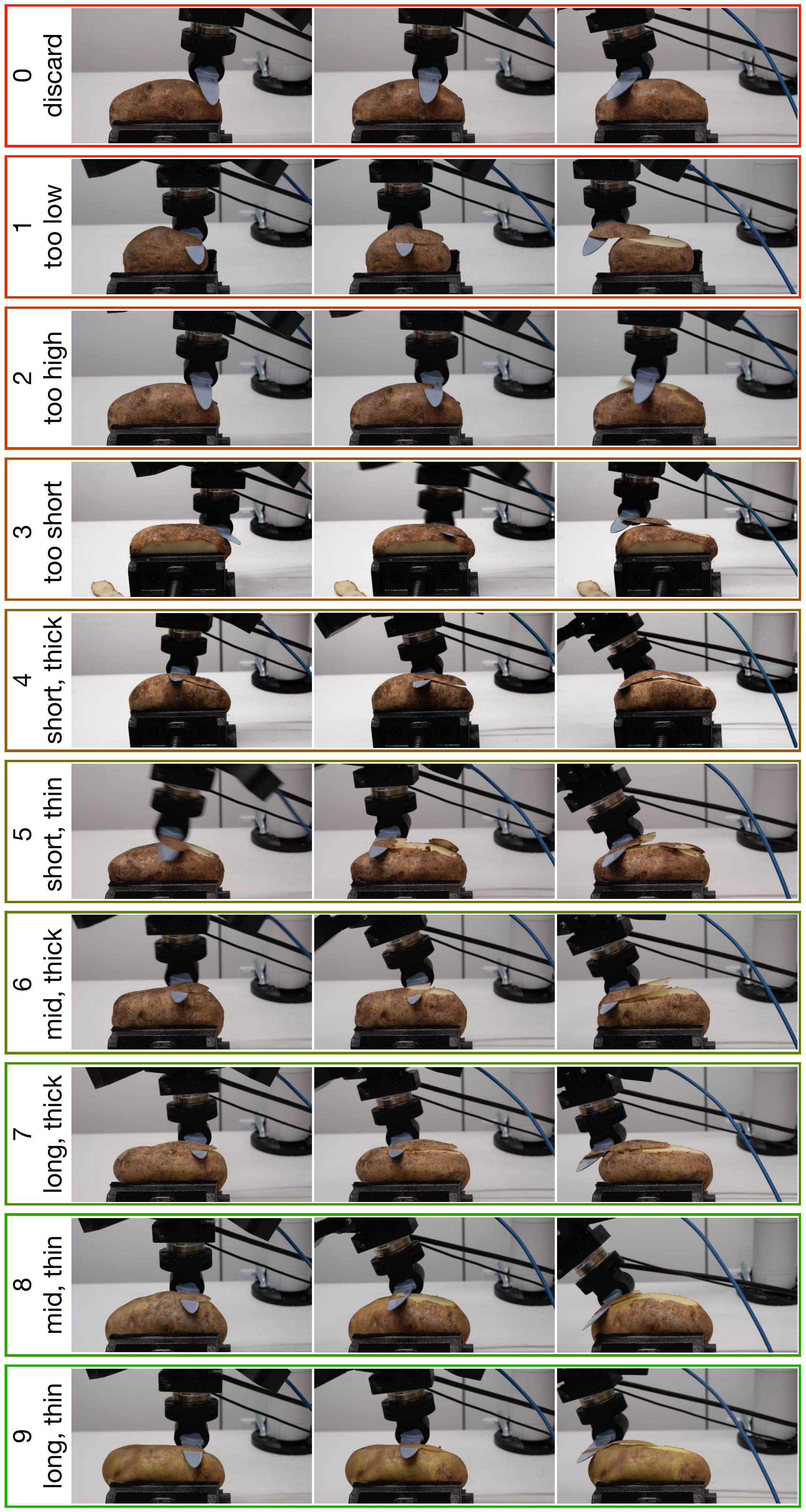}
    \caption{\small\textbf{Front-view visualization of qualitative score metric.}
   We use integer scores from 0 to 9 (the higher the better) to capture subjective human preferences based on the overall visual appearance of the peel.}
    \label{fig:qual}
\end{figure}
\begin{figure}[h!]
    \centering
    \includegraphics[width=0.55\linewidth]{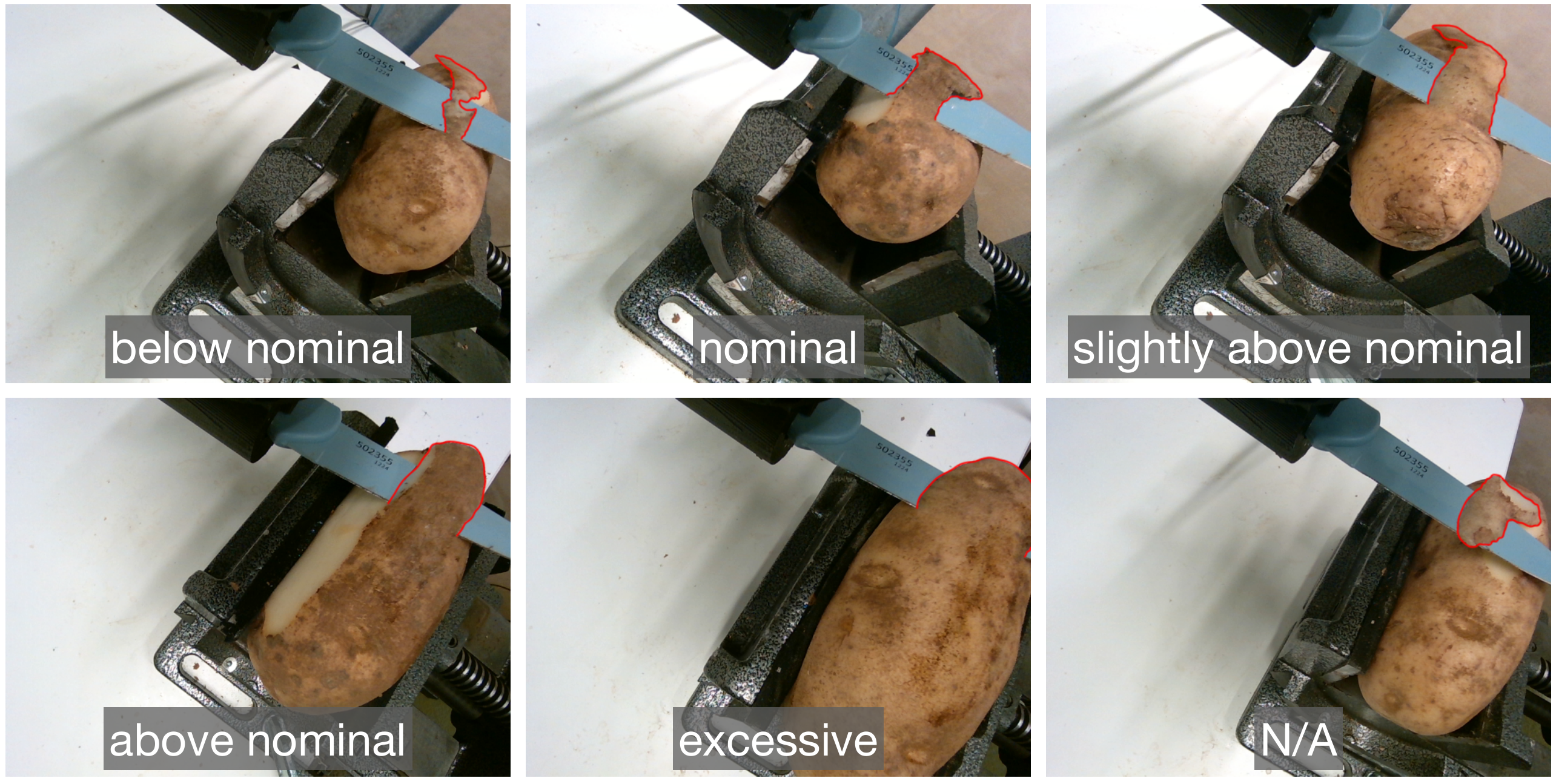}
    \caption{\small\textbf{Wrist-view visualization of quantitative score metric.} We use six discrete thickness categories, where \textit{nominal} denotes the most desired thickness. Details of how these categories are mapped to normalized scalar rewards can be found in Appendix~\ref{app:reward}.}
    \label{fig:quant}
\end{figure}
\FloatBarrier

\section{Compliant Controller Details}
\label{app:compliance}

We execute policy commands using a torque-level joint-space impedance controller with internal compliance adaptation.
The controller runs at $\Delta t = 0.002\,\mathrm{s}$ (500\,Hz) and outputs joint torques at each control step.
All gain matrices ($K_p, K_d, K_r, K_l, K_{lp}$) are diagonal and positive definite unless otherwise specified.

Given desired joint trajectories $(q_d^t, \dot q_d^t)$ generated by an online trajectory generator and sensed joint states $(q_s^t, \dot q_s^t, \tau_s^t)$, we first compute a task tracking torque
\begin{equation}
\tau_{\text{task}}^{t}
= -K_p\big(q_n^{t}-q_d^{t}\big)
  -K_d\big(\dot q_n^{t}-\dot q_d^{t}\big)
  + g(q_s^{t}),
\end{equation}
where $q_n^t, \dot q_n^t$ denote internal nominal joint states and $g(q_s^t)$ is the gravity compensation torque computed from the sensed joint configuration.

To induce compliance under external contact, the nominal joint acceleration is updated according to the discrepancy between the commanded torque and the measured joint torque, $\ddot q_n^{t}=K_r^{-1}\big(\tau_{\text{task}}^{t} - \tau_s^{f,t}\big)$, where $K_r$ is a diagonal stiffness matrix and $\tau_s^{f,t}$ is a low-pass filtered version of the sensed torque, $\tau_s^{f,t}=\alpha\,\tau_s^{t} + (1-\alpha)\,\tau_s^{f,t-1}$.
This formulation allows the nominal trajectory to adapt when sustained torque discrepancies are observed, effectively introducing admittance-like compliance while retaining torque-level control.

The nominal joint velocity and position are updated using semi-implicit Euler integration, $\dot q_n^{t+1}=\dot q_n^{t} + \ddot q_n^{t}\,\Delta t$ and $q_n^{t+1}=q_n^{t} + \dot q_n^{t+1}\,\Delta t$.

To reduce drift between nominal and sensed trajectories during prolonged contact, we apply a friction-like coupling term
\begin{equation}
\tau_f^{t}
= K_r K_l\Big( (\dot q_n^{t}-\dot q_s^{t})
+ K_{lp}(q_n^{t}-q_s^{t}) \Big),
\end{equation}
which damps relative motion and improves stability by softly pulling the nominal state toward the sensed state.

The final commanded torque is $\tau_c^{t} = \tau_{\text{task}}^{t} + \tau_f^{t}$.

Based on this formulation, the controller parameters used in our experiments are listed in Table~\ref{table:compliance}.
\begin{table}[h]
\renewcommand\arraystretch{0.95}
\centering
\begin{tabular*}{0.65\linewidth}{l@{\extracolsep{\fill}}c}
\toprule
$\alpha$  & 0.01 \\
$K_r$            & [0.3, 0.3, 0.3, 0.3, 0.18, 0.18, 0.18]    \\
$K_l$              & [106.2, 100.8, 106.2, 106.2, 131.4, 106.2, 106.2]     \\
$K_{lp}$              & [11.89, 25.52, 22.0, 22.0, 22.0, 22.0, 22.0]     \\
$K_{p}$              & [382.2, 296.4, 347.1, 400.0, 200.0, 200.0, 200.0]     \\
$K_{d}$              & [21.0, 17.5, 10.0, 10.0, 5.0, 5.0, 5.0]     \\
\bottomrule
\end{tabular*}
\caption{Impedance controller parameters.}
\label{table:compliance}
\end{table}

\section{Reward Design Details}
\label{app:reward}

We use a hybrid reward formulation that combines \textit{quantitative} and \textit{qualitative} components. Each demonstration is annotated at two temporal resolutions: a segment-level quantitative score measuring relative peeling thickness, and a trajectory-level qualitative score capturing overall execution preference.
Both label-to-reward mappings are structured lookup tables ordered by physical desirability rather than free hyperparameters: the quantitative reward peaks at the nominal thickness and decreases as the peel deviates from it in either direction, and the qualitative reward follows the same monotonic ordinal rubric for all produce. Apple and potato differ only in the quantitative reward scale (Table~\ref{tab:quant_reward_map}) and the reward weight $\alpha$ (Section~\ref{app:combined}), because a given absolute thickness is penalized more on thin-skinned apples than on potatoes.

\subsection{Quantitative reward}
Quantitative scores are provided at the segment level (2\,Hz). We use six discrete thickness categories (Figure~\ref{fig:quant}): \textit{below nominal}, \textit{nominal}, \textit{slightly above nominal}, \textit{above nominal}, \textit{excessive}, and \textit{N/A}. These categories are mapped to normalized scalar rewards $\mathcal{R}_{\mathrm{quant}} \in [0,1]$ via task-specific lookup tables for apple and potato (Table~\ref{tab:quant_reward_map}).

\begin{table}[h]
\centering
\setlength{\tabcolsep}{7pt}
\begin{tabular}{lcc}
\toprule
Quantitative label & Apple $\mathcal{R}_{\mathrm{quant}}$ & Potato $\mathcal{R}_{\mathrm{quant}}$ \\
\midrule
below nominal            & 0.3 & 0.5 \\
nominal                  & 1.0 & 1.0 \\
slightly above nominal   & 0.8 & 0.5 \\
above nominal            & 0.3 & 0.1 \\
excessive                & 0.0 & 0.0 \\
N/A                      & 0.0 & 0.0 \\
\bottomrule
\end{tabular}
\caption{Quantitative reward mapping for apple and potato.}
\label{tab:quant_reward_map}
\end{table}

Quantitative rewards are converted into per-step signals by uniformly assigning the segment reward to all frames within the segment. To reduce boundary discontinuities, we apply a lightweight linear smoothing over the $O$ overlapping frame pairs between adjacent segments. For the $i$-th overlap ($i=0,\dots,O-1$), we interpolate as $\alpha_i=(i+1)/(O+1)$ and $r \leftarrow (1-\alpha_i)\, r_{\mathrm{prev}} + \alpha_i\, r_{\mathrm{next}}$, then assign the interpolated value symmetrically to both sides of the segment boundary.

\subsection{Qualitative reward}
Each trajectory additionally receives a single qualitative preference score in the range $[0,9]$ (Figure~\ref{fig:qual}), reflecting overall execution quality such as consistency, smoothness, and perceived preference. The qualitative score is mapped to a normalized scalar reward $\mathcal{R}_{\mathrm{qual}} \in [0,1]$ using the task-specific lookup table shown in Table~\ref{tab:global_reward_map}.

\begin{table}[h]
\centering
\setlength{\tabcolsep}{7pt}
\begin{tabular}{cccc}
\toprule
Qualitative score & Descriptor & Apple $\mathcal{R}_{\mathrm{qual}}$ & Potato $\mathcal{R}_{\mathrm{qual}}$ \\
\midrule
0 & discard & 0.0 & 0.0 \\
1 & too low & 0.1 & 0.1 \\
2 & too high & 0.2 & 0.2 \\
3 & too short & 0.3 & 0.3 \\
4 & short, thick & 0.4 & 0.4 \\
5 & short, thin & 0.5 & 0.5 \\
6 & mid, thick & 0.6 & 0.6 \\
7 & long, thick & 0.8 & 0.8 \\
8 & mid, thin & 0.9 & 0.9 \\
9 & long, thin & 1.0 & 1.0 \\
\bottomrule
\end{tabular}
\caption{Qualitative reward mapping for apple and potato.}
\label{tab:global_reward_map}
\end{table}

\subsection{Combined reward}
\label{app:combined}
At each timestep, the final reward is computed by combining quantitative and qualitative components:
\begin{equation}
r =
\begin{cases}
\mathcal{R}_{\mathrm{quant}}, & \text{if } \mathcal{R}_{\mathrm{quant}} < \tau, \\
\alpha\, \mathcal{R}_{\mathrm{quant}} + (1-\alpha)\, \mathcal{R}_{\mathrm{qual}}, & \text{otherwise}.
\end{cases}
\end{equation}

In all experiments, we set $\tau=0.1$ and use segment length $L=15$ with overlap $O=3$. The weighting parameter $\alpha$ is set to $0.85$ for the apple task and $0.75$ for the potato task.

\section{IK Solver Details}
\label{app:ik}

To produce smoother end-effector motion in Cartesian space, we implement a weighted least-squares inverse kinematics (IK) solver that dynamically adjusts how much each joint moves. Instead of treating all joints equally, it assigns adaptive weights computed from the Jacobian to penalize joints that cause large leverage, favoring distal joints (elbow/wrist) over proximal ones (base/shoulder). Our solver also uses a weighted null-space projector to stay close to the default pose without reintroducing jitter or violating the smoothness constraint.

\section{Training Details and Hyperparameters}
\label{app:training}

\paragraphc{Training details} To train the final policies, we collect 50, 150, and 200 demonstrations for cucumber, apple, and potato, respectively, with each demonstration consisting of a single peel stroke; ablation studies in Appendix~\ref{app:learning} use the numbers of trajectories specified there. All RGB-D observations are resized to $(120,160,4)$ in height, width, and channels. Force-torque measurements from the ATI sensor are represented as a 6-dimensional vector, and proprioceptive input consists of a 7-dimensional joint-angle vector. The visual encoder is a ResNet-18 with a 64-dimensional output. The state encoder is a two-layer feedforward network with hidden dimension 64 and output dimension 64. We apply dropout with rate 0.1 to mitigate overfitting. For the diffusion policy, we use a two-layer feedforward denoising network with hidden dimension 64, which predicts 6-dimensional end-effector actions.

\section{Data Collection Method Details}
\label{app:datacollection}

\begin{figure}[htbp]
    \centering
    \includegraphics[width=0.55\linewidth]{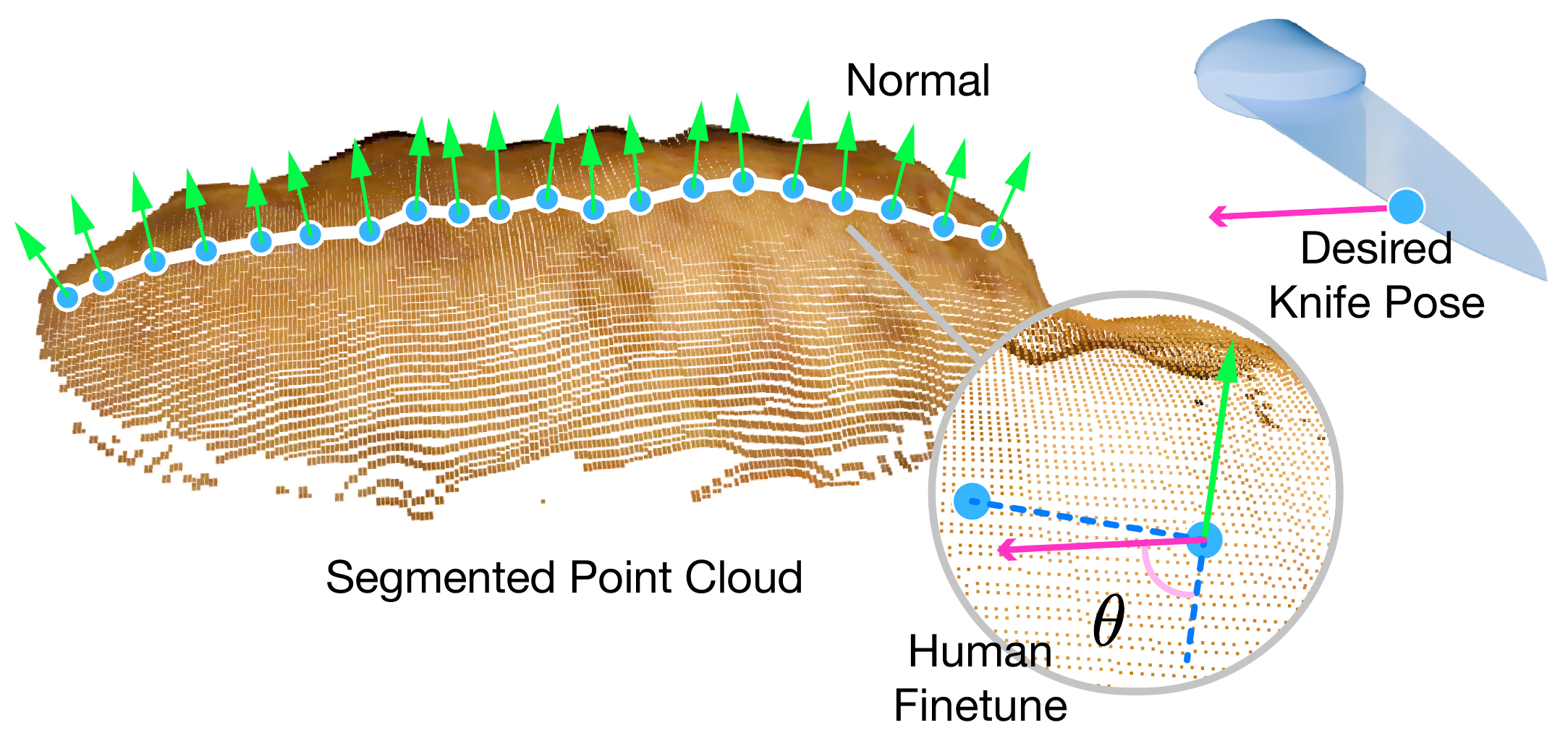}
    \caption{\small\textbf{Model-based planner.} A visualization of the planning procedure of our heuristic planner. We use a calibrated third-view LiDAR camera to capture the scene, extract a segmented point cloud of the object, and plan a dense trajectory on its surface that represents the peeling path. We then calculate the desired knife pose trajectory based on the surface normal of each waypoint on the path, and execute the planned trajectory with human-in-the-loop correction of knife pose via keyboard control (e.g. +20 degrees in yaw).}
    \label{fig:planner}
\end{figure}

\paragraphc{Model-based planner} We use a calibrated third-view L515 LiDAR camera to capture RGBD images of the peeling scene, and implement a planner based on visual inputs. Specifically, we first extract a segmented point cloud of the object, plan a dense trajectory on its surface with 20 waypoints, extract surface normal of the object at each waypoint, calculate the desired knife pose trajectory, and solve for the desired joint trajectory using IK. A visualization of the planning procedure is shown in Figure~\ref{fig:planner}.

\paragraphc{VR teleoperation} The \uline{VR teleoperation} baseline is implemented following prior work~\cite{lin2024learning}. We find that it performs much worse than SpaceMouse teleoperation on this peeling task. Reasonably, the instability of the physical action of holding the VR controllers and the noise in VR tracking performance both prevent teleoperator from reaching the precision and delicacy required for peeling.

\paragraphc{Kinesthetic teaching} We evaluate \uline{kinesthetic teaching} in two stages: first, collection of position trajectories; second, trajectory replay with hand-tuned compliance parameters using heuristics similar to ACP~\cite{hou2025adaptive}. This is for fairer comparison with other methods, since trajectories collected with kinesthetic teaching have different effective compliance parameters from the original ones used by controller and cannot be simply recovered. From the results, we draw three findings: (1) it is substantially faster to collect peeling data with kinesthetic teaching; (2) its raw data quality is only mildly lower than that with SpaceMouse (Table~\ref{table:data}) -- likely because guiding the arm by hand requires strenuous yet precise muscle control; (3) however, replaying the collected trajectories fails entirely, even after extensive parameter tuning, because the effective compliance parameters during collection cannot be recovered for autonomous execution. Therefore, despite being faster, kinesthetic teaching does not yield usable training data in our setup, and we use SpaceMouse teleoperation instead.

\section{Observation and Action Space Design Details}
\label{app:learning}

\begin{figure}[h]
    \centering
    \includegraphics[width=0.32\linewidth]{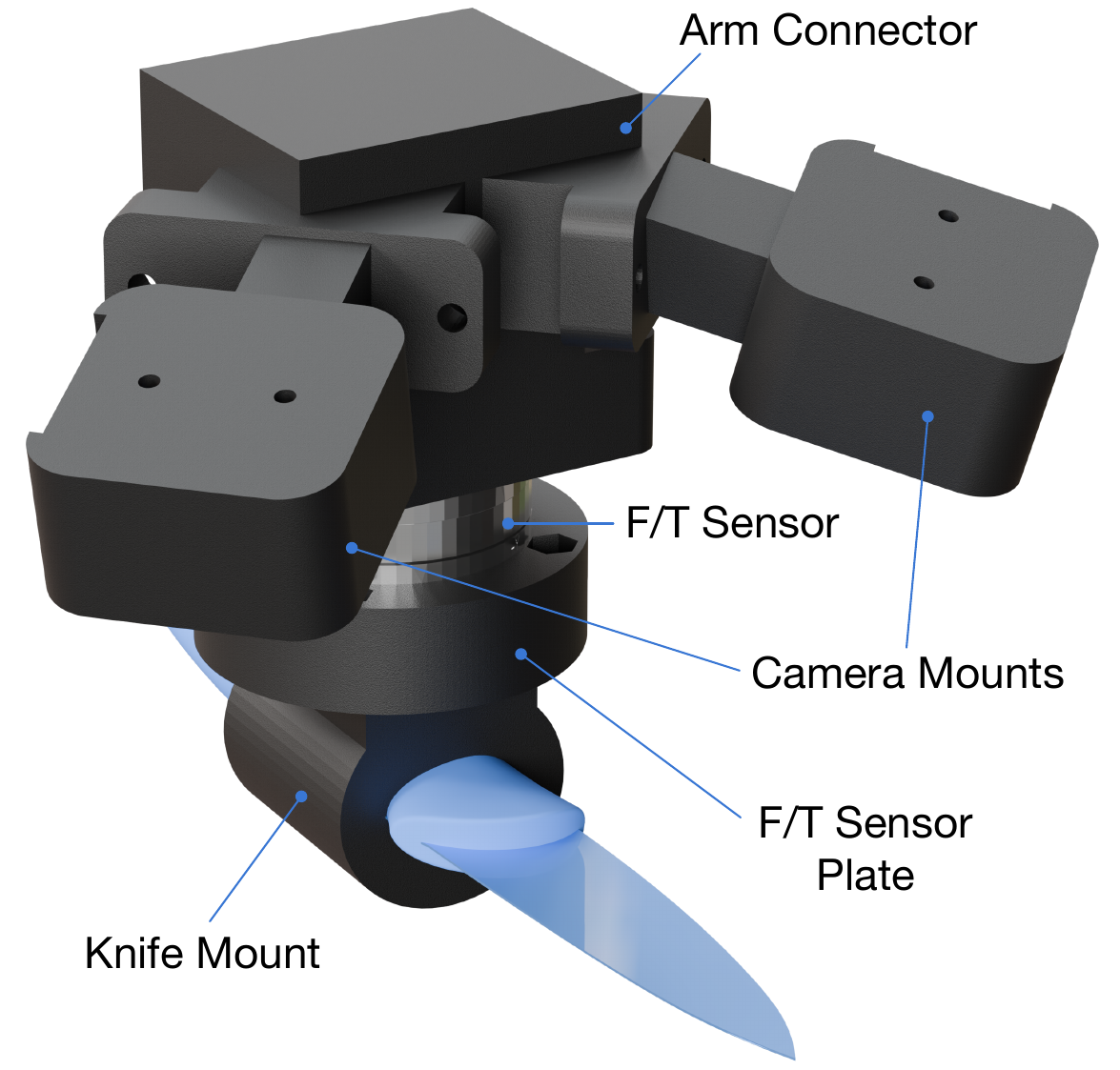}
    \caption{\small\textbf{End-effector mount.} A CAD visualization of our custom end-effector mount design, including an arm connector, a force-torque sensor plate, two camera mounts, and a knife mount.}
    \label{fig:mount}
\end{figure}

\begin{table}[h!]
\centering
\hfill
\adjustbox{valign=t}{%
\begin{minipage}{.31\linewidth}
\centering
\captionsetup{justification=raggedright, singlelinecheck=off}
\caption{\textbf{Wrist camera.} We study the relative importance of two wrist cameras on potato data. Interestingly, the \textit{before} camera contributes to policy performance more than the \textit{after} camera, suggesting the benefit of learning from a less occluded view.}
\setlength{\tabcolsep}{0.8em}
\renewcommand{\arraystretch}{1.2}
\resizebox{0.98\linewidth}{!}{%
\begin{tabular}{lcc}
\toprule
{Num Cam} & {Success \%} & {Avg. Score} \\
\cmidrule(r){1-1}
\cmidrule(l){2-3}
Both & 80\ci{38}{96} & 5.9 \\
\textit{Before} Only & 100\ci{57}{100} & 4.8 \\
\textit{After} Only & 60\ci{23}{88} & 3.2 \\
\bottomrule
\end{tabular}
}
\label{tab:wrist_cam}
\end{minipage}
}%
\hfill
\adjustbox{valign=t}{%
\begin{minipage}{.31\linewidth}
\centering
\captionsetup{justification=raggedright, singlelinecheck=off}
\caption{\textbf{Data modalities.} We denote grayscale RGB images as gRGB, original RGB images as RGB, depth images as D, and force-torque readings as F. Visual observation with grayscaled RGB and force-torque observation are both important.}
\setlength{\tabcolsep}{1.3em}
\renewcommand{\arraystretch}{1.2}
\resizebox{0.98\linewidth}{!}{%
\begin{tabular}{lcc}
\toprule
{Modality} &  {Success \%} & {Avg. Score} \\
\cmidrule(r){1-1}
\cmidrule(l){2-3}
gRGB, D, F & 80\ci{38}{96} & 5.9 \\
gRGB, D & 60\ci{23}{88} & 5.2 \\
RGB, D, F & 0\ci{0}{43} & 1.6 \\
F & 0\ci{0}{43} & 0.6 \\
\bottomrule
\end{tabular}
}
\label{tab:modalities}
\end{minipage}
}%
\hfill
\adjustbox{valign=t}{%
\begin{minipage}{.31\linewidth}
\centering
\captionsetup{justification=raggedright, singlelinecheck=off}
\caption{\textbf{Sample efficiency.} We study the correlation between number of trajectories and policy performance on potato data. We find that performance improves almost linearly in terms of both success rate and qualitative score as number of trajectories increases until success rate reaches 80\%.}
\setlength{\tabcolsep}{1.3em}
\renewcommand{\arraystretch}{1.3}
\resizebox{0.98\linewidth}{!}{%
\begin{tabular}{lcc}
\toprule
{Num Traj} & {Success \%} & {Avg. Score} \\
\cmidrule(r){1-1}
\cmidrule(l){2-3}
200 (100\%) & 80\ci{49}{94} & 5.9 \\
100 (50\%) & 60\ci{31}{83} & 4.4 \\
40 (20\%) & 10\ci{2}{40} & 2.4 \\
\bottomrule
\end{tabular}
}
\label{tab:efficiency}
\end{minipage}
}%
\hfill
\end{table}

\FloatBarrier

\paragraphc{Designing observation and action spaces} We compare important design choices in camera placement, number of cameras, and choice of data modalities that enable precise, adaptive and generalizable peeling. For each design choice, we perform ablation experiments by comparing the task success rate and generalization. Results for each policy are obtained from running 5 evaluation trials with the best performing checkpoint on potatoes.
\begin{itemize}
    \item Wrist camera: As shown in Figure~\ref{fig:mount}, we attach two wrist cameras to the end effector mount. Subtly, since our peeling direction is fixed, one camera always capture the object and knife view slightly \textit{before} the current peeling action, while the other camera always capture the object and knife view slightly \textit{after}. In Table~\ref{tab:wrist_cam}, we present an ablation study comparing using both cameras, only the \textit{before} camera, and only the \textit{after} camera. Using both cameras yields the highest average score (5.9, vs.\ 4.8 and 3.2 for a single camera), although the \textit{before}-only policy succeeds in more of the 5 trials; this is reasonable since two cameras provide more information, including implicit 3D information. Furthermore, we find that the \textit{before} camera contributes to policy performance more than the \textit{after} camera. We hypothesize this is because the \textit{before} camera captures a less occluded view of the contact between knife edge and produce than the \textit{after} camera, due to the produce's curved geometry.

    \item Data modalities: Our policy takes three types of sensing modalities as inputs: proprioception, vision, and touch. For vision, we use RGBD images where the original colored RGB is converted to grayscale. In Table~\ref{tab:modalities}, we quantitatively investigate how the input modalities affect policy performance. Results show that it is important to (1) use both visual and force-torque observations, and (2) convert RGB to grayscale. We reason that (1) is because of the force-sensitive and position-sensitive nature of our peeling task, and (2) because grayscale image input forces the policy to focus on geometric features rather than produce texture, greatly aiding generalization.

\end{itemize}

\paragraphc{Sample efficiency} We study the efficiency of learning by empirically evaluating the correlation between number of demonstrations and policy performance on potato. The results are shown in Table~\ref{tab:efficiency}. For potato peeling, 200 trajectories are needed to reach 80\% success rate, with an average score of 5.9. This amounts to about 33 potatoes, where each potato contributes about 6 trajectories. Similarly, with 50 trajectories (about 8 cucumbers) our cucumber peeling policy achieves 100\% success rate and an average score of 8.5; with 100 trajectories  (about 17 apples) our apple peeling policy achieves 60\% success rate and an average score of 3.8. For each reported number, the evaluation is conducted over 10 trials.

\end{document}